\documentclass[letterpaper]{article} 
\usepackage{aaai25}  
\usepackage{times}  
\usepackage{helvet}  
\usepackage{courier}  
\usepackage[hyphens]{url}  
\usepackage{graphicx} 
\usepackage{natbib}  
\usepackage{caption} 
\usepackage{algorithm}
\usepackage{algorithmic}

\usepackage{mathtools}
\usepackage{amssymb}
\usepackage{amsthm}
\usepackage{multirow}
\usepackage{subfigure}
\usepackage{booktabs}
\usepackage{newfloat}
\usepackage{listings}
\DeclareCaptionStyle{ruled}{labelfont=normalfont,labelsep=colon,strut=off} 
\floatstyle{ruled}
\newfloat{listing}{tb}{lst}{}
\floatname{listing}{Listing}
\title{Large Language Models Are Read/Write Policy-Makers for \\ Simultaneous Generation}
\author{
    Shoutao Guo \textsuperscript{\rm 1,3},
    Shaolei Zhang \textsuperscript{\rm 1,3},
    Zhengrui Ma \textsuperscript{\rm 1,3},
    Yang Feng \textsuperscript{\rm 1,2,3}\thanks{ Corresponding author: Yang Feng.}
}
\affiliations{
        \textsuperscript{\rm 1}{Key Laboratory of Intelligent Information Processing,} \\ Institute of Computing Technology, Chinese Academy of Sciences (ICT/CAS) \\
    { \textsuperscript{\rm 2} {Key Laboratory of AI Safety, Chinese Academy of Sciences}} \\
    { \textsuperscript{\rm 3} {University of Chinese Academy of Sciences, Beijing, China}}

    guoshoutao22z@ict.ac.cn, zhangshaolei20z@ict.ac.cn, fengyang@ict.ac.cn
}

\usepackage{bibentry}

\begin{document}

\maketitle

\begin{abstract}
Simultaneous generation models write generation results while reading streaming inputs, necessitating a policy-maker to determine the appropriate output timing. Existing simultaneous generation methods generally adopt the traditional encoder-decoder architecture and learn the generation and policy-making capabilities through complex dynamic programming techniques. Although LLMs excel at text generation, they face challenges in taking on the role of policy-makers through traditional training methods, limiting their exploration in simultaneous generation. To overcome these limitations, we propose a novel LLM-driven Simultaneous Generation (LSG) framework, which allows the off-the-shelf LLM to decide the generation timing and produce output concurrently. Specifically, LSG selects the generation policy that minimizes latency as the baseline policy. Referring to the baseline policy, LSG enables the LLM to devise an improved generation policy that better balances latency and generation quality, and writes generation results accordingly. Experiments on simultaneous translation and streaming automatic speech recognition tasks show that our method can achieve state-of-the-art performance utilizing the open-source LLMs and demonstrate practicality in real-world scenarios.
\end{abstract}

%
\begin{links}
    \link{Code}{https://github.com/ictnlp/LSG}
\end{links}

\section{Introduction}
Simultaneous generation models \citep{reinforcement, streaming_asr}, which produce the target sentence before reading the entire input, are widely used in streaming scenarios such as real-time subtitles and online meetings. To achieve the goal of low latency and high generation quality \citep{DualPath}, simultaneous generation models require an optimal policy to determine the generation timing, ensuring that the generated results are consistent with those in non-streaming scenarios while minimizing latency \citep{alinejad-etal-2021-translation}. Consequently, the learning of generation policy is critical to the simultaneous generation tasks.

In simultaneous generation tasks such as simultaneous translation \citep{waitk} and streaming Automatic Speech Recognition (ASR) \citep{streaming_asr}, existing methods are constrained to using non-streaming parallel data for model training due to the lack of annotated policies. To learn the generation policy, previous methods \citep{DBLP:conf/iclr/MaPCPG20, DBLP:conf/emnlp/MiaoBS21} primarily utilize an encoder-decoder architecture \citep{DBLP:conf/nips/VaswaniSPUJGKP17} coupled with complex dynamic programming training techniques. This methodology endows simultaneous generation models with both generation and policy-making capabilities \citep{DBLP:journals/corr/abs-2303-00257}. However, these models are constrained by their expressive capacity, resulting in suboptimal policies and generation performance. Additionally, they suffer from significant memory consumption and slow training speeds during training \citep{guo-etal-2023-learning}. More recently, the emergence of Large Language Models (LLMs) \citep{touvron2023llama2openfoundation} prompts researchers to explore their potential in simultaneous generation tasks \citep{koshkin2024transllamallmbasedsimultaneoustranslation, agostinelli2024simulllmframeworkexploringhighquality}.
\begin{figure}[t]
    \centering
    \includegraphics[width=3.15in]{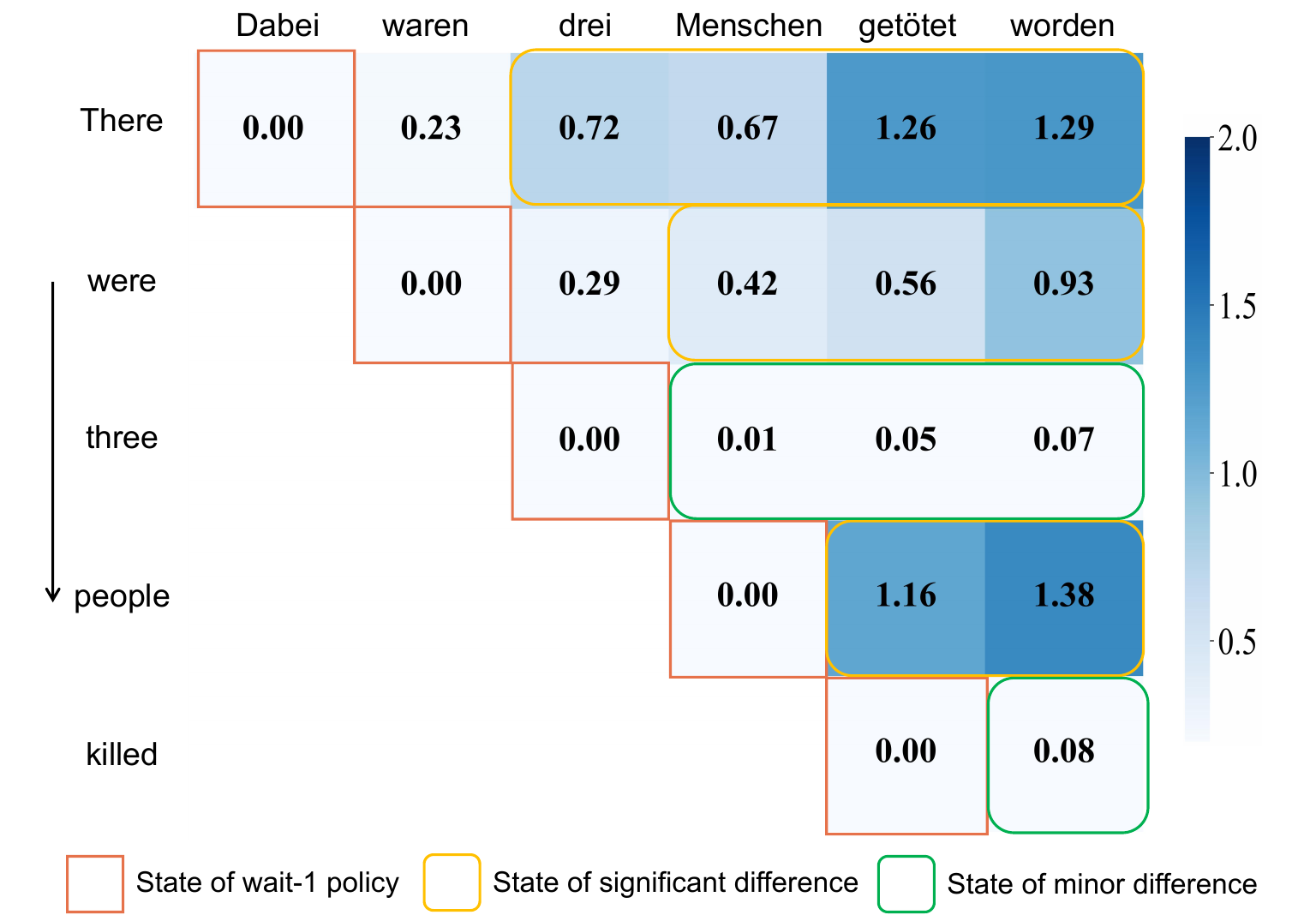}
    \caption{The distribution difference of subsequent generation states compared to wait-1 policy for a German$\Rightarrow$English translation example. The distribution difference is measured by KL divergence.}
    \label{pre_analysis}
\end{figure}
Nevertheless, the decoder-only architecture and vast parameters of LLMs pose challenges in applying traditional dynamic programming methods for policy learning. Consequently, existing LLM-based methods leverage the generation capabilities of LLMs to produce outputs guided by either fixed policies \citep{waitk} or policies provided by conventional encoder-decoder models \citep{guo2024agentsimtagentassistedsimultaneousmachine}. 
Unfortunately, these external policies not only introduce complex control processes but also result in inferior performance without considering the context of LLMs. Therefore, incorporating LLMs into simultaneous generation tasks remains challenging.

To bypass the need for policy training and derive effective policies for LLMs, a straightforward approach might be to compare the current outputs with the non-streaming results, generating the target words only when the two align. This is akin to deriving a new policy from a full-sentence policy, where the model can use the complete input for generation. However, this is not feasible in practice, as the model cannot access the entire input in advance. On the other hand, minimum source input is available during simultaneous generation. This insight leads us to consider whether we can derive a policy by comparing the generation results based on minimum input with those based on the current input.

Therefore, we attempt to develop an enhanced policy that improves upon a \emph{baseline policy}, which defines the minimum input at each generation step. To validate our hypothesis, we conduct a comprehensive preliminary analysis. We utilize the wait-1 policy \citep{waitk} as the baseline policy and \texttt{Llama2-7B-chat} \citep{touvron2023llama2openfoundation} as the LLM. Initially, we leverage the LLM to obtain the generation distribution for target words at each generation state, based on available source content. We then analyze the distribution differences between the baseline policy and subsequent generation states. Figure \ref{pre_analysis} illustrates a notable trend where the distribution differences gradually increase as more source content is processed. Crucially, once the necessary source content is available, the distribution differences become significant, indicating an opportune moment for generation. These findings suggest that leveraging distribution differences can effectively strike trade-offs between latency and generation quality. However, Figure \ref{pre_analysis} also highlights a special case where all distribution differences of some target words remain relatively minor, as the wait-1 policy already provides sufficient information for generation. This phenomenon, inherently influenced by language characteristics and word reordering, is unavoidable and necessitates specialized treatment in our approach.


In light of these insights, we propose the LLM-driven Simultaneous Generation (LSG) method, a novel approach that empowers the off-the-shelf LLM to determine the policies and generate outputs concurrently. Our LSG method enables the LLM to derive an enhanced policy from a baseline policy without needing policy learning. At each step, the LLM compares the distribution difference between the current input and the source content determined by the baseline policy. When this distribution difference reaches a predetermined threshold, the LLM is prompted to generate outputs. Otherwise, LSG continues to await the upcoming input. To address the special case illustrated in Figure \ref{pre_analysis}, we utilize the confidence of the LLM to avoid excessive delays that might be caused by minor distribution differences. To validate the effectiveness of LSG, we conduct extensive experiments on simultaneous translation and streaming ASR tasks. Leveraging open-source LLMs, our method achieves state-of-the-art results on standard datasets and demonstrates practicality in real-world scenarios.

\section{Background}
\label{background_tag}
\paragraph{Simultaneous Generation} 
Let $\mathbf{x}$ = $(x_1, ..., x_J)$ denote the complete source sequence, where $x_i$ represents a source word or a speech segment. The simultaneous generation model incrementally produces the target sentence $\mathbf{y}$ = $(y_1, ..., y_I)$ with length $I$ based on a generation policy. To represent this policy, we introduce the notation $g_i$, which represents the length of the partial input sequence when generating $y_i$. Therefore, the policy for generating $\mathbf{y}$ from the source sequence $\mathbf{x}$ can be defined as $\mathbf{g}$ = $(g_1, ..., g_I)$. During inference, the simultaneous generation model generates the target sentence according to the following formula:
\begin{equation}
    p(\mathbf{y}|\mathbf{x}, \mathbf{g}) = \sum\limits_{i=1}^{I} p(y_i \mid \mathbf{x}_{\leq g_i}, \mathbf{y}_{<i}),
\end{equation}
where $p(y_i \mid \mathbf{x}_{\leq g_i}, \mathbf{y}_{<i})$ is the next token distribution.

\paragraph{Wait-k policy}
\label{waitk_des}
Simultaneous generation models require a policy to determine the timing of generating sentences. Currently, the most prevalent simultaneous generation policy is the wait-k policy \cite{waitk}, which is simple and exhibits relatively inferior performance. During inference, the wait-k policy initially reads $k$ source elements (i.e., speech segments or words), then alternates between generating a word and reading a source element. Therefore, the wait-k policy can be expressed by the following equation:
\begin{equation}
    g^{wait-k}_i = \min \bigl\{ k+i-1, J \bigr\},
\end{equation}
where $J$ denotes the length of the whole input sequence. According to the Average Lagging metric \citep{waitk} for latency evaluation, the policy with the minimum latency is the wait-0 policy. However, the wait-0 policy is impractical, as it would result in the simultaneous generation model producing the first word without conditioning on any source information. Therefore, we select the wait-1 policy as the baseline policy for our method.

\begin{figure*}[t]
    \centering
    \includegraphics[width=6.0in]{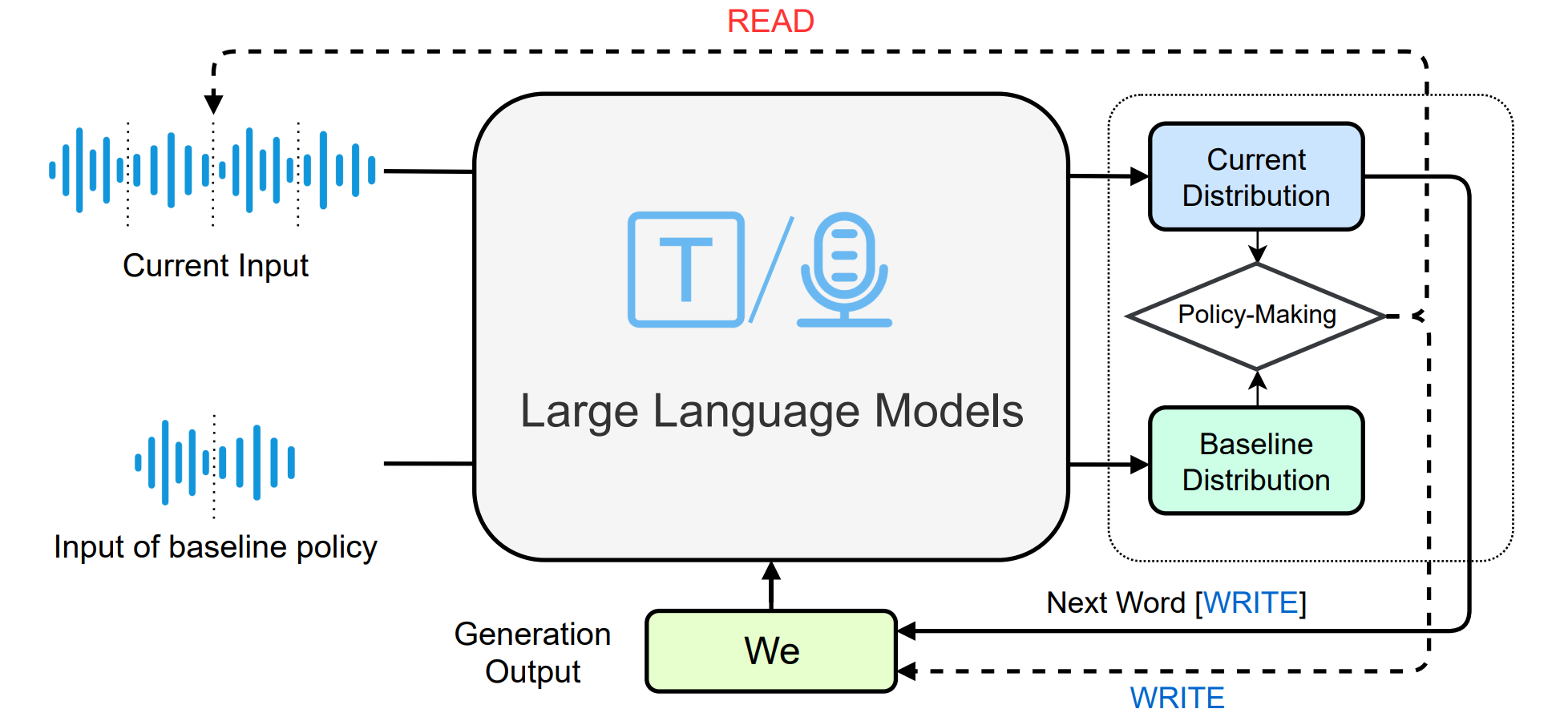}
    \caption{The framework of LLM-driven Simultaneous Generation Model.}
    \label{fig-model}
\end{figure*}

\section{Method}
In this section, we introduce our LLM-driven Simultaneous Generation (LSG) method, which empowers the LLM to perform policy-making and generation sub-tasks concurrently. We first present the framework of LSG and delineate its operational process. Subsequently, we elucidate how the LLM leverages a baseline policy to derive an enhanced policy. To address the limitations of the baseline policy in scenarios where the source content is already sufficient, we introduce an additional confidence condition for the enhanced policy. Finally, we implement a range constraint for the obtained policy to ensure controllable latency and mitigate the impact of some outlier policies. The following subsections provide a detailed exposition of our method.

\subsection{Model Framework}
As shown in Figure \ref{fig-model}, we introduce the model framework of LSG. Our LSG method empowers the LLM to perform both policy-making and generation sub-tasks. To this end, the LLM pre-establishes a baseline policy before initiating simultaneous generation.

At each generation step, LSG selects the source content corresponding to the baseline policy as a new input, based on the currently available input and previously generated words. Subsequently, LSG enables the LLM to predict the next target word based on the current input and the input determined by the baseline policy, respectively. This process yields two probability distributions: the current generation distribution and the distribution of baseline policy. LSG utilizes these distributions for policy-making to determine the action to be taken. If the READ action is selected, LSG refrains from producing any output at that moment and awaits the upcoming input. Conversely, if the WRITE action is chosen, LSG generates the target word based on the current distribution and appends it to the previously generated words. After that, a new generation step commences.

Our framework does not impose restrictions on the employed LLMs. However, the baseline policy needs to be predetermined in advance of simultaneous generation. As discussed in the Background section, to ensure low latency and usability of the baseline policy, we choose the wait-1 policy as our baseline policy.

\subsection{Policy-Making Procedure}
In this subsection, we elaborate on the policy-making procedure in Figure \ref{fig-model}. Our LSG method aims to develop an improved policy by referencing the baseline policy. At each generation step, it utilizes the differences between the current generation distribution with the distribution of the wait-1 policy to decide on the taken action.

At the current generation step, we assume that the available source sequence is $\mathbf{x}_{\leq j}$ and generated target words are $\mathbf{y}_{<i}$, where $j$ is greater than $i$. Therefore, we can obtain $p(y_i | \mathbf{x}_{\leq j}, \mathbf{y}_{<i})$, which denotes the generation distribution of the LLM based on $\mathbf{x}_{\leq j}$ and $\mathbf{y}_{<i}$. At the same time, under the guidance of the wait-1 policy, the LLM utilizes $\mathbf{x}_{\leq i}$ and $\mathbf{y}_{< i}$ to generate the distribution $p(y_i | \mathbf{x}_{\leq i}, \mathbf{y}_{<i})$. These two distributions are used by LSG to calculate the KL divergence to decide on the action to be taken:
\begin{equation}
    \label{contrastive}
    \mathbb{D}_{\textrm{KL}}\bigl[ p(y_i | \mathbf{x}_{\leq j}, \mathbf{y}_{<i})\mid \mid p(y_i | \mathbf{x}_{\leq i}, \mathbf{y}_{<i}) \bigr] > \delta,
\end{equation}
where $\delta$ is the hyperparameter that represents the threshold. If the condition in Eq.(\ref{contrastive}) is met, LSG generates the target word based on the distribution $p(y_i | \mathbf{x}_{\leq j}, \mathbf{y}_{<i})$ and appends it to the previously generated sequence. Otherwise, our method refrains from producing any output and waits for the upcoming input. 

\paragraph{Confidence Condition}
Up to now, we have developed improved policies by referencing the baseline policy without needing traditional complex training methods \citep{DBLP:journals/corr/abs-2303-00257}. However, due to factors such as language features and word reordering \citep{liu-etal-2021-cross}, the baseline policy may have already provided sufficient source information for some target words. As illustrated in Figure \ref{pre_analysis}, this phenomenon can result in minor distribution differences when generating these words according to the condition in Eq.(\ref{contrastive}). We call this phenomenon as \emph{false negative}, as it instructs the model to excessively read source information even if condition in Eq.(\ref{contrastive}) is met, resulting in redundant latency \citep{papi-etal-2023-attention}. However, this phenomenon is unavoidable due to the diversity of language expression. To complement the condition in Eq.(\ref{contrastive}), we introduce an additional confidence condition.

Since LLMs typically assign probability mass to favorable behaviors \citep{li-etal-2023-contrastive}, the confidence of LLMs also reflects the credibility of the generation. In the face of the false negative problem in the condition of Eq.(\ref{contrastive}), we use the confidence of LLMs to mitigate this issue:
\begin{equation}
    \label{confidence}
    \max p(y_i | \mathbf{x}_{\leq j}, \mathbf{y}_{<i}) > \alpha,
\end{equation}
where $\alpha$ is the confidence hyperparameter that enables generation. Consequently, our LSG method executes the WRITE action when either the condition in Eq.(\ref{contrastive}) or Eq.(\ref{confidence}) is satisfied. Otherwise, LSG awaits the upcoming input.

\subsection{Range constraint}
After introducing the policy-making procedure, our LSG method can leverage the LLM to perform both policy-making and generation sub-tasks. However, when considering the practical applications, there are still issues with the above policy-making procedure. In the current setup, the search range for the target word $y_i$ is $[\min \bigl\{i, J\bigr\}, J]$, where $J$ denotes the length of the whole input sequence. However, the presence of outlier policies will inevitably lead to excessive latency or poor translation quality \citep{DBLP:conf/iclr/MaPCPG20}. Moreover, it is challenging to ensure that the simultaneous generation model always responds within a fixed delay. Therefore, it is necessary to impose constraints on the search range of the policy.

In our LSG method, we set the search range for the target word $y_i$ as:
\begin{equation}
    [\min \bigl\{L+i-1, J\bigr\}, \min \bigl\{L+i-1+U, J\bigr\} ],
\end{equation}
where $L$ denotes the number of pre-read elements before simultaneous generation and $U$ represents the degree of autonomy afforded to the LLM in policy-making.

\section{Experiments}

\subsection{Datasets}
We mainly conduct experiments on simultaneous text-to-text translation (SimulT2TT), simultaneous speech-to-text translation (SimulS2TT), and streaming ASR tasks. 

\textbf{WMT15\footnote{\url{www.statmt.org/wmt15}} German$\Rightarrow$English (De$\Rightarrow$En)} We conduct SimulT2TT task on this dataset. Consistent with \citet{DBLP:conf/iclr/MaPCPG20}, we use the newstest2015 set as the test set.

\textbf{MuST-C English$\Rightarrow$German (En$\Rightarrow$De)} This dataset \citep{di-gangi-etal-2019-must} is collected from TED talks and we conduct the SimulT2TT task using its text data.

\textbf{CoVoST2 French$\Rightarrow$English (Fr$\Rightarrow$En)} We use this dataset \citep{wang2020covost2massivelymultilingual} to conduct both SimulS2TT and streaming ASR tasks.

\subsection{System Settings}
Since our method can be applied to SimulT2TT, SimulS2TT, and streaming ASR tasks, we will delineate the comparative methods for each of these tasks separately and then present the settings of our LSG method.

For SimulT2TT task, the baseline methods include \textbf{wait-k} \citep{waitk}, \textbf{MMA} \citep{DBLP:conf/iclr/MaPCPG20}, \textbf{ITST} \citep{ITST}, \textbf{HMT} \citep{DBLP:journals/corr/abs-2303-00257} and \textbf{Agent-SiMT} \citep{guo2024agentsimtagentassistedsimultaneousmachine}. With the exception of Agent-SiMT, the aforementioned methods all use the traditional encoder-decoder architecture. HMT, which learns policies through sophisticated dynamic programming training methods, achieves the superior performance among conventional approaches. Agent-SiMT, leveraging an agent collaboration mechanism and utilizing policies provided by HMT to guide the LLMs in translation generation, has achieved state-of-the-art performance in the SimulT2TT task.

For SimulS2TT task, we compare our method against \textbf{DiSeg} \citep{zhang-feng-2023-end} and \textbf{StreamSpeech} \citep{zhang2024streamspeechsimultaneousspeechtospeechtranslation}. Both DiSeg and StreamSpeech adopt the encoder-decoder architecture, with StreamSpeech achieving state-of-the-art performance in the SimulS2TT task. To validate the practical applicability of our method, we additionally evaluate all approaches using computation-aware latency metrics for this task.

For streaming ASR task, \textbf{Wav2Vec2-large} \citep{baevski2020wav2vec20frameworkselfsupervised} and \textbf{Whisper-base} \citep{radford2022robustspeechrecognitionlargescale} are used as the baseline methods. Both Wav2Vec2 and Whisper are pre-trained models, with Whisper demonstrating superior performance across multiple ASR datasets.

Since LSG is a general simultaneous generation framework, it does not impose restrictions on the LLMs used. Due to the constraints of different tasks, we employ different LLMs for different evaluated tasks. For the SimulT2TT task, we maintain the same setup as \citet{guo2024agentsimtagentassistedsimultaneousmachine}. We employ \texttt{Llama2-7B-chat}\footnote{\url{https://huggingface.co/meta-llama/Llama-2-7b-chat-hf}} as the LLM and perform fine-tuning on 10w extracted samples using LoRA \citep{hu2021lora}. For the SimulS2TT and streaming ASR tasks, we use the open-source speech LLM, Qwen-Audio\footnote{\url{https://github.com/QwenLM/Qwen-Audio}} \citep{chu2023qwenaudioadvancinguniversalaudio}. As the multimodal version of the Qwen \citep{bai2023qwentechnicalreport} series, Qwen-Audio achieves good comprehension and generation capabilities in multiple speech tasks after audio-language pre-training. During inference, the duration of each speech segment is set to 640 ms. The prompt templates used in our experiments are consistent with those used during the training of the LLMs. We set $\delta$ = 9.0 and $\alpha$ = 0.6 for De$\Rightarrow$En task, $\delta$ = 7.5 and $\alpha$ = 0.6 for En$\Rightarrow$De task, and $\delta$ = 7.0 and $\alpha$ = 0.5 for Fr$\Rightarrow$En task. For different latency scenarios, we set [$L$, $U$] as [1, 4], [3, 4], [5, 6], and [7, 6], respectively.


\subsection{Evaluation}
In evaluating streaming generation systems, we employ the SimulEval toolkit \citep{ma-etal-2020-simuleval} to assess two critical aspects: latency and generation quality. Systems that demonstrate low latency while maintaining high generation quality are generally considered superior.

To quantify latency, we utilize the Average Lagging (AL) metric \citep{waitk}, which measures the delay between input reception and output generation in simultaneous generation systems. For textual input, AL is calculated in terms of word count, whereas for speech input, it is measured in milliseconds (ms). Additionally, for the SimulS2TT task, we evaluate computation-aware latency on an NVIDIA RTX 3090 GPU, which assesses the latency of the systems in practical applications.

To assess generation quality, we employ task-specific metrics. For SimulT2TT and SimulS2TT tasks, we utilize the SacreBLEU metric \citep{post-2018-call}, a widely used metric in translation. For the streaming ASR task, we adopt the Word Error Rate (WER) as our primary evaluation metric.

\begin{figure*}[t]
\centering
\subfigure[WMT15 De$\Rightarrow$En]
{
\label{main_1}
\includegraphics[width=2.18in]{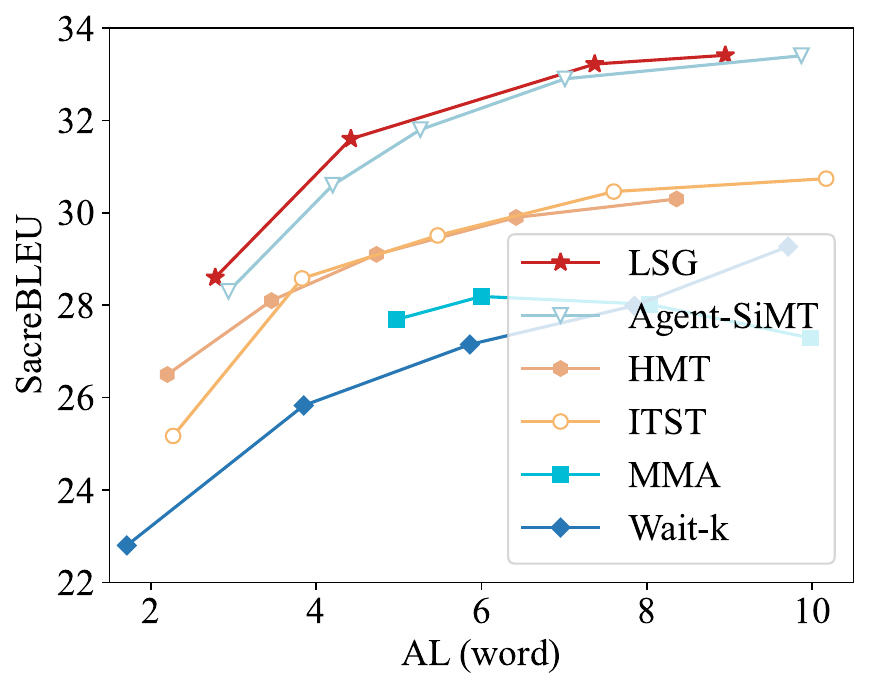}
}\hspace{0.1cm}
\subfigure[MuST-C En$\Rightarrow$De]{
\label{main_2}
\includegraphics[width=2.18in]{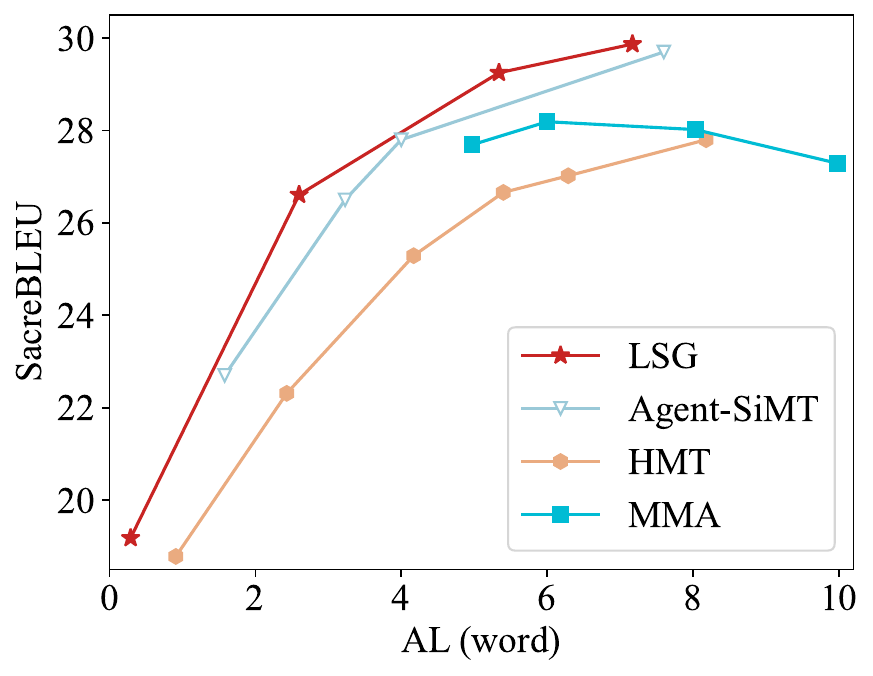}
}\hspace{0.1cm}
\subfigure[CoVoST2 Fr$\Rightarrow$En]{
\label{main_3}
\includegraphics[width=2.18in]{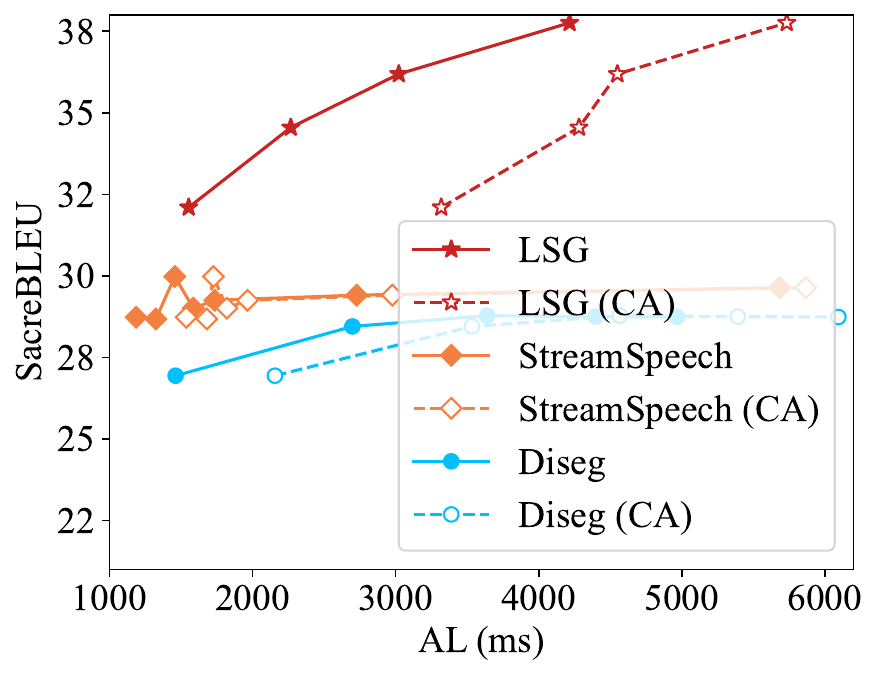}
}

\caption{Performance of simultaneous generation models on De$\Rightarrow$En, En$\Rightarrow$De and Fr$\Rightarrow$En datasets. We also evaluate the Computation-Aware (CA) latency on the CoVoST2 Fr$\Rightarrow$En dataset to assess the usability of systems in real-world scenarios.}
\label{main_res}
\end{figure*}

\begin{table}[]
\centering
\begin{tabular}{c c  c} \toprule[1.2pt]
\textbf{Method} & \textbf{AL (ms) ($\downarrow$)} & \textbf{WER ($\downarrow$)}                                       \\ \cmidrule(lr){1-1} \cmidrule(lr){2-3}

\textbf{Wav2Vec2-large} & 5684.38 &  26.17 \\

\textbf{Whisper-base} & 5684.38 &  38.04 \\

\cmidrule(lr){1-1} \cmidrule(lr){2-3}

\multirow{2}{*}{\textbf{LSG}} & 3161.25 & 31.71 \\

                        & 4342.23 & 23.76 \\

 \bottomrule[1pt]
\end{tabular}
\caption{The streaming ASR performance of simultaneous generation models on the CoVoST 2 Fr$\Rightarrow$En dataset.}
\label{asr}
\end{table}

\subsection{Main Results}
We evaluate the performance of our method on SimulT2TT, SimulS2TT, and streaming ASR tasks.

For the SimulT2TT task, we present the performance of various simultaneous generation models in Figure \ref{main_1} and Figure \ref{main_2}. Our method achieves state-of-the-art performance across both datasets. Compared to traditional approaches \citep{DBLP:conf/iclr/MaPCPG20, DBLP:journals/corr/abs-2303-00257} that utilize the encoder-decoder framework, our method demonstrates significant improvements in simultaneous translation performance. Conventional methods require the design of intricate policy modules integrated into the transformer architecture \citep{DBLP:conf/nips/VaswaniSPUJGKP17}, followed by training through sophisticated dynamic programming techniques. However, these traditional methods are often constrained by their expressive capacity, resulting in inferior generation performance. Our approach leverages the enhanced comprehension and generation capabilities of LLMs, leading to superior performance. In addition to the traditional methods, our method also outperforms LLM-based methods \citep{guo2024agentsimtagentassistedsimultaneousmachine}. Previous LLM-based methods necessitate coupling an external policy module with the LLM to accomplish simultaneous translation tasks, which fails to provide appropriate policies for the LLM and increases system complexity. In contrast, our method allows LLMs to utilize their inherent understanding capabilities to acquire policies, which then guide the translation generation process. This results in better trade-offs between latency and translation quality.

For the SimulS2TT task, Figure \ref{main_3} compares our method with other simultaneous speech translation methods. As the first method to utilize LLMs for simultaneous speech translation, our approach outperforms previous methods across all latency levels. Previous approaches rely on speech pre-training models \citep{zhang-feng-2023-end}, multi-task training \citep{zhang2024streamspeechsimultaneousspeechtospeechtranslation}, and dynamic programming strategies \citep{liu-etal-2021-cross} to enhance performance. However, these methods necessitate complex and multiple training processes and are constrained by the generation capabilities of the model. In contrast, our method transforms off-the-shelf speech LLMs into simultaneous speech translation systems directly, serving both policy-making and generation roles. By leveraging the speech understanding and instruction-following capabilities of Qwen-Audio, our method significantly further improves simultaneous speech translation performance. Additionally, we provide results for computation-aware latency, which considers both the delay between input and output and the model inference time, reflecting the latency of real-world scenarios. Despite using speech LLMs, our method can respond to speech input with a delay of only 3 seconds, demonstrating its practical applicability. Moreover, our method can be accelerated with better GPUs and inference frameworks, making it well-suited for simultaneous speech translation tasks.

For the streaming ASR task, we compare our method with previous pre-trained speech models, as shown in Table \ref{asr}. Our LSG method achieves recognition quality comparable to previous methods with a delay of 6 seconds while maintaining only about a delay of 3 seconds. Although the methods based on pre-trained models have been trained on large amounts of speech data, they often lack language generation capabilities and struggle to establish effective generation policies. In contrast, by utilizing the speech comprehension and language generation abilities of speech LLMs \citep{chu2023qwenaudioadvancinguniversalaudio}, our approach provides superior generation policies in streaming scenarios. By combining advantages in both generation and policy, our method achieves better streaming ASR performance.

Therefore, by leveraging the policy-making and generation capabilities of off-the-shelf LLMs, our LSG method can attain the best generation performance across multiple simultaneous generation tasks. 

\section{Analysis}
To deepen the understanding of our approach, we conduct extensive analyses. We then introduce each analytical experiment in detail separately.

\subsection{Ablation Study}

\begin{table}[]
\centering
\begin{tabular}{l c  c} \toprule[1.2pt]
\textbf{$\;\;\;\;\;\;\;$Method} & \textbf{AL (word) ($\downarrow$)} & \textbf{SacreBLEU ($\uparrow$)}  \\ 

\cmidrule(lr){1-1} \cmidrule(lr){2-3}

\multirow{2}{*}{\textbf{LSG} }& 4.42 &  31.60 \\
                            & 7.37 &  33.22 \\

\cmidrule(lr){1-1} \cmidrule(lr){2-3}

\multirow{2}{*}{$\;\;$\textbf{w/o Confidence}} & 4.89 &  31.34 \\
                              & 6.75 &  32.72 \\
                              
\cmidrule(lr){1-1} \cmidrule(lr){2-3}

\multirow{2}{*}{$\;\;$\textbf{w/o Range}} & 3.62 & 21.95 \\

                        & 12.91 & 29.90 \\

 \bottomrule[1pt]
\end{tabular}
\caption{The ablation experiments of our method, where ``w/o Confidence" represents the removal of the confidence condition and ``w/o Range" indicates our method without range constraint. The experimental results are all based on the De$\Rightarrow$En task.}
\label{ablation_study}
\end{table}

\begin{table}[]
\centering
\begin{tabular}{c c  c} \toprule[1.2pt]
\textbf{Segment Size (ms)} & \textbf{AL (ms) ($\downarrow$)} & \textbf{SacreBLEU ($\uparrow$)}  \\

\cmidrule(lr){1-1} \cmidrule(lr){2-3}

\multirow{2}{*}{\textbf{320}} & 1566.42 &  31.71 \\
                              & 3003.99 & 36.08 \\

\cmidrule(lr){1-1} \cmidrule(lr){2-3}

\multirow{2}{*}{\textbf{640}} & 1582.94 &  32.20 \\
             & 3022.18 &  36.19 \\

\cmidrule(lr){1-1} \cmidrule(lr){2-3}

\textbf{960} & 3101.12 &  36.47 \\

 \bottomrule[1pt]
\end{tabular}
\caption{Ablation study on speech segment size in the SimulS2TT task. The experimental results are based on the Fr$\Rightarrow$En dataset.}
\label{segment_size}
\end{table}

To explore the impact of different settings in our method, we conduct several ablation experiments. 

Table \ref{ablation_study} demonstrates that all components of our LSG method contribute to the performance of simultaneous generation. Firstly, the introduction of the confidence condition mitigates the false negative problem inherent in using only the condition in Eq.(\ref{contrastive}). This confidence condition enables our method to select the WRITE action when the current generation does not satisfy the condition in Eq.(\ref{contrastive}) but exhibits high confidence. This allows our method to avoid unnecessary delays caused by waiting for additional source information \citep{tang-etal-2023-hybrid}, consequently achieving superior performance. More importantly, the range constraint facilitates even more substantial improvements in our method. By employing this constraint, our approach effectively controls the scope and autonomy of LLMs in determining generation policies. This constraint allows us to limit the policy-making range of LLMs based on linguistic features \citep{DBLP:conf/emnlp/MiaoBS21}, striking better trade-offs while ensuring timely responses.

We also investigate the influence of segment size when processing speech input. Table \ref{segment_size} illustrates the performance of our method on the SimulS2TT task across various segment sizes. The results indicate that our approach exhibits robustness to changes in source speech segment size. While a segment size of 960 achieves relatively strong performance, it lacks the flexibility to adapt to low-latency requirements in practical applications. Conversely, a segment size of 320 necessitates more frequent LLM inferences, resulting in increased computational costs. Consequently, we opt for a speech segment size of 640 in our experimental setup. This choice delivers superior performance among the three configurations while allowing for flexible latency adjustments to meet diverse operational needs.

\begin{figure}[t]
    \centering
    \includegraphics[width=3.15in]{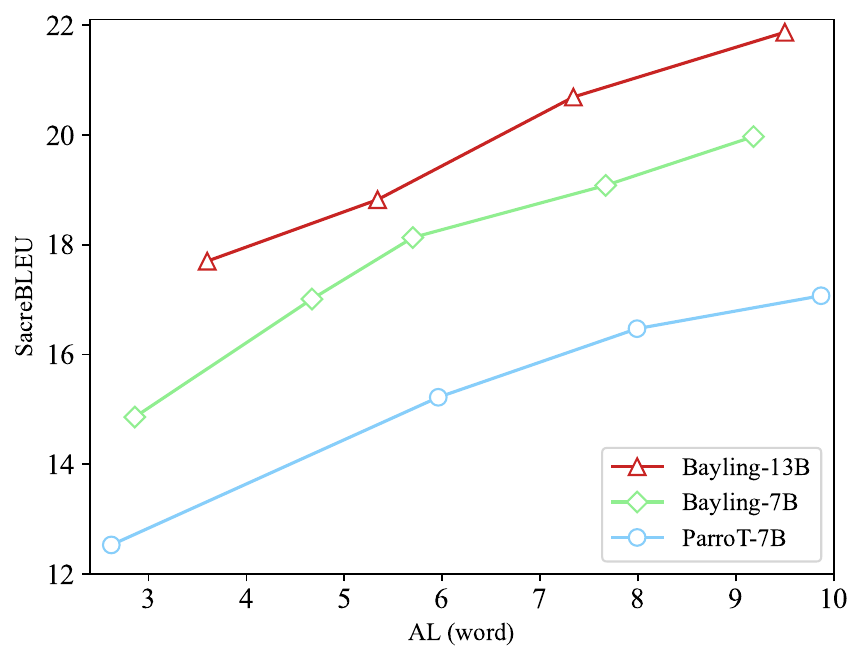}
    \caption{The performance of LSG framework when employing various LLMs. The results are reported on the WMT22 Chinese$\Rightarrow$English dataset.}
    \label{wmt22_zhen}
\end{figure}

\begin{table}[]
\centering
\begin{tabular}{c | c c c } \toprule[1.2pt]
\textbf{LLMs} & ParroT-7B & Bayling-7B & Bayling-13B  \\

\cmidrule(lr){1-1} \cmidrule(lr){2-4}

\textbf{SacreBLEU} & 18.73 &  20.72 & 23.57 \\

 \bottomrule[1pt]
\end{tabular}
\caption{The performance of LLMs in non-streaming scenarios. The numerical results are based on the WMT22 Chinese$\Rightarrow$English dataset.}
\label{zhen_full}
\end{table}

\subsection{Influence of LLMs}
Following our ablation experiments, we further analyze the impact of different LLMs on simultaneous generation performance. Our objective is to investigate whether more advanced LLMs can yield better simultaneous generation results within our LSG framework. 

To this end, we evaluate ParroT-7B \citep{jiao-etal-2023-parrot}, Bayling-7B \citep{zhang2023bayling}, and Bayling-13B on the WMT22\footnote{\url{https://www.statmt.org/wmt22}} Chinese$\Rightarrow$English translation dataset. We initially assess the performance of these LLMs in non-streaming scenarios in Table \ref{zhen_full}. The results demonstrate that the models of the Bayling family outperform ParroT-7B, achieving superior translation quality. Moreover, Bayling-13B, with its advantages of more parameters, surpasses the performance of Bayling-7B.

Building upon the insights of non-streaming performance, we then integrate these LLMs into our LSG framework. Figure \ref{wmt22_zhen} illustrates the performance of our method when utilizing different LLMs. Leveraging their enhanced Chinese$\Rightarrow$English translation capabilities, the models of the Bayling family achieve better trade-offs between latency and translation quality. Notably, Bayling-13B, with its substantial number of parameters, attains superior performance in simultaneous translation compared to Bayling-7B.

These findings underscore that our method serves as a versatile, unified framework applicable to existing LLMs. Furthermore, it demonstrates the potential to achieve enhanced streaming generation performance when integrated with more advanced LLMs.

\begin{figure}[t]
    \centering
    \includegraphics[width=3.15in]{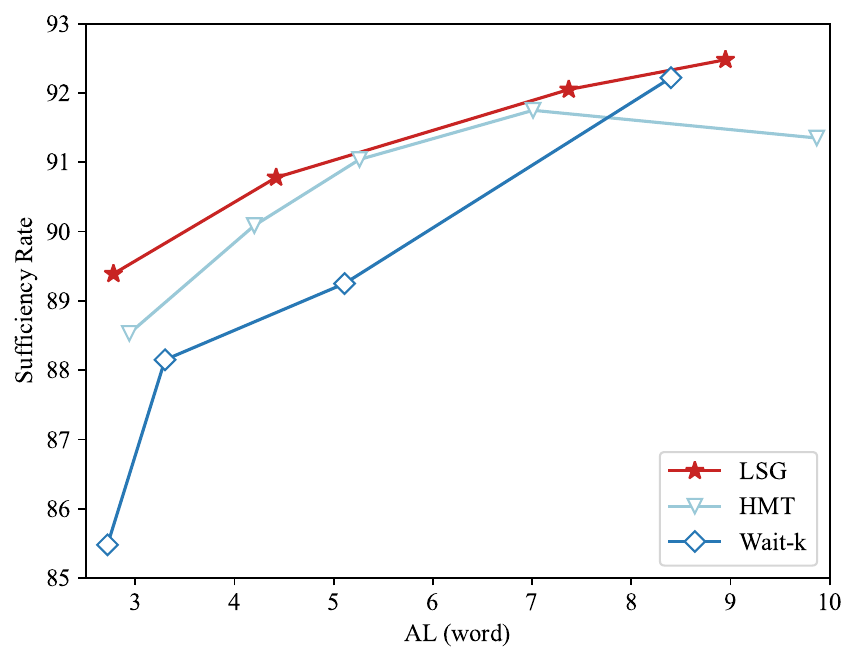}
    \caption{Comparison of the policy sufficiency of different simultaneous generation policies. The experiments are based on the De$\Rightarrow$En dataset.}
    \label{suff}
\end{figure}

\subsection{Quality of Policy}

After exploring the relationship between LLMs and simultaneous generation performance, we further investigate the quality of the policies obtained by our LSG method. In simultaneous generation, generation is considered sufficient if the target word is produced after reading the aligned source information under the guidance of policy \citep{guo2024decoderonlystreamingtransformersimultaneous}. Conversely, when LLMs rely solely on their anticipation capabilities for next-token prediction, the outcome is undesired. Therefore, we want to compare the sufficiency of the generation outputs under different policies to validate the quality of our learned policy.

To this end, we employ the eflomal\footnote{\url{https://github.com/robertostling/eflomal}} toolkit to obtain input-output alignments and calculate generation sufficiency. We evaluate the sufficiency of our LSG method against external policies such as wait-k and HMT when applied to the Llama2-7B-chat \citep{touvron2023llama2openfoundation} model. The results in Figure \ref{suff} show that our method consistently achieves higher generation sufficiency under all latency. Leveraging the comprehension capabilities of LLMs, our method enables the LLM to develop superior policies, surpassing the sufficiency of generation under the guidance of external policies. This underscores that our LSG method empowers LLMs to acquire suitable policies without the need for explicit policy learning.

\section{Related Work}
\textbf{SimulT2TT} Recent SimulS2TT methods are broadly divided into two categories: encoder-decoder and LLMs. The approaches using the encoder-decoder architecture initially employ the wait-k policy \citep{waitk} and enhance performance through training methods \citep{elbayad2020efficient, chen-etal-2021-improving-simultaneous, guo-etal-2023-simultaneous, 10446517}. Further efforts in this line of work employ techniques such as monotonic attention \citep{arivazhagan-etal-2019-monotonic, DBLP:conf/iclr/MaPCPG20}, wait-info \citep{zhang-etal-2022-wait}, hidden Markov models \citep{DBLP:journals/corr/abs-2303-00257}, CTC-based non-autoregressive structure \citep{ma-etal-2023-non, ma-etal-2024-non} to conduct policy learning and translation concurrently. With the advent of LLMs, some methods \citep{ agostinelli2024simulllmframeworkexploringhighquality} attempt to utilize external policy to guide LLMs.

\textbf{SimulS2TT} Recent SimulS2TT approaches mainly focus on adapting speech segmentation or enhancing model structures. Initial method \citep{ma2020simulmtsimulstadaptingsimultaneous} attempts to split source speech into fixed-length segments. Subsequent work tries to adaptively segment speech using techniques such as auxiliary ASR task \citep{zeng-etal-2021-realtrans, chen-etal-2021-direct}, integrate-and-fire model \citep{dong-etal-2022-learning}, and differentiable segmentation \citep{zhang-feng-2023-end}, applying the wait-k policy to the resulting segments. Other work enhances SimulS2TT performance through enhanced architectures such as augmented Transducer \citep{liu-etal-2021-cross} and combinations of transducer and encoder-decoder model \citep{tang-etal-2023-hybrid, ma2024learningmonotonicattentiontransducer}. To the best of our knowledge, no prior research has explored the potential of leveraging LLMs to address the SimulS2TT task.

\textbf{Streaming ASR} Previous Streaming ASR methods primarily rely on transducer \citep{yeh2019transformertransducerendtoendspeechrecognition, 9054715} and attention-based \citep{fan19b_interspeech, 8683510} architectures. More recently, the robust performance of pre-trained speech models \citep{baevski2020wav2vec20frameworkselfsupervised, radford2022robustspeechrecognitionlargescale} in various ASR tasks has also led to their widespread adoption in streaming ASR tasks.

Previous simultaneous generation methods rarely explore the use of LLMs and cannot fully harness the policy-making and generation capabilities of LLMs. Therefore, our LSG method enables the off-the-shelf LLM to develop improved policies by considering a baseline policy and then completing generation accordingly. This allows the LLM to autonomously and efficiently complete the simultaneous generation without the need for complex training methods.

\section{Conclusion}

In this paper, we propose a novel LLM-driven simultaneous generation method that allows the LLMs to decide the generation timing and produce output concurrently. Experiments show that our method achieves state-of-the-art performance demonstrates practicality in real-world scenarios.

\section{Acknowledgments}
This paper is supported by National Natural Science Foundation of China (Grant No. 62376260). We thank all the anonymous reviewers for their valuable feedback and thorough reviews.

\bibliography{aaai2025_arxiv}

\section{Appendix}

\subsection{Sufficiency Rate}

In simultaneous generation, generation is deemed sufficient when the target word is produced after reading the aligned source information under the guidance of the policy \citep{yu2024selfmodifyingstatemodelingsimultaneous}. Consequently, the sufficiency rate under different policies serves as a crucial indicator of policy quality, as demonstrated in our analysis section. Here, we provide a detailed explanation of the methodology for calculating the sufficiency rate.

Given a source sequence $\mathbf{x} = (x_1, ..., x_J)$, the LLM can incrementally generate the translation $\mathbf{y} = (y_1, ..., y_I)$ guided by the policy $\mathbf{g} = (g_1, ..., g_I)$. 

Utilizing the alignment tool, we can determine the input-output alignments and subsequently obtain the sequence $\mathbf{a} = (a_1, ..., a_I)$, where $a_i$ denotes the index of the last source element aligned with target word $y_i$. The sufficiency rate is then calculated as follows:
\begin{equation}
    S = \frac{1}{I} \sum\limits_{i=1}^{I} \mathbb{I}(a_i \leq g_i),
\end{equation}
where $\mathbb{I}(\cdot)$ is the indicator function, returning 1 when the condition $a_i \leq g_i$ is satisfied and 0 otherwise. This formulation allows us to quantify the sufficiency of generation, thereby providing a robust metric for evaluating policy quality in simultaneous generation tasks.

\subsection{Case Study}
To provide a more comprehensive understanding of our LSG method, we present a case analysis. We select two examples from the CoVoST2 Fr$\Rightarrow$En test set, specifically focusing on the SimulS2TT task.

\begin{figure*}[t]
    \centering
    \includegraphics[width=6.0in]{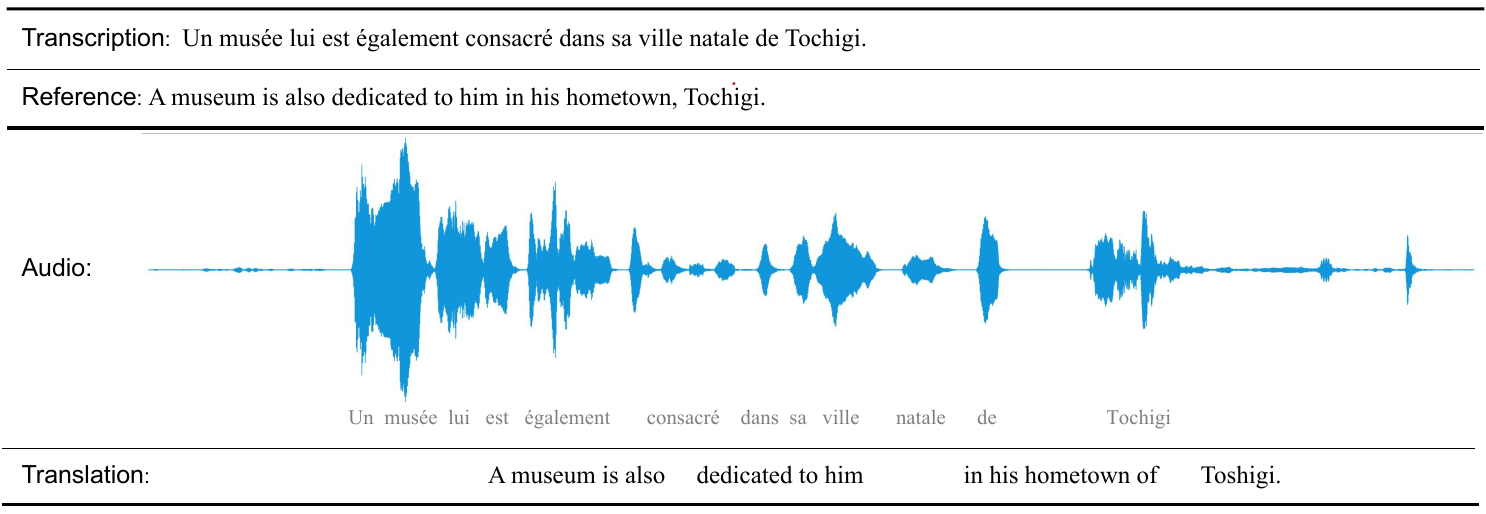}
    \caption{The results of LSG method on common\_voice\_fr\_19617179 from the CoVoST2 Fr-En test set.}
    \label{case1}
\end{figure*}

\begin{figure*}[t]
    \centering
    \includegraphics[width=6.0in]{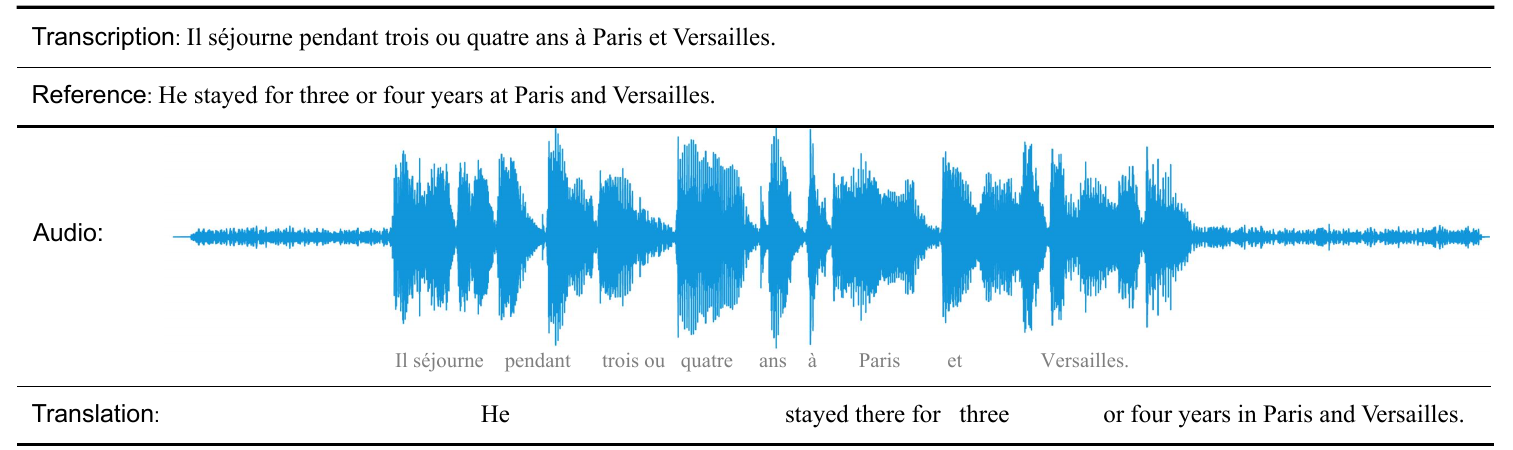}
    \caption{The results of LSG method on common\_voice\_fr\_19439076 from the CoVoST2 Fr-En test set.}
    \label{case2}
\end{figure*}

Figures \ref{case1} and \ref{case2} illustrate our analysis. In these figures, `Transcription' refers to the French text corresponding to the audio, while `Reference' denotes the ground-truth English translation of the French text. As the audio sequence is processed, LSG demonstrates its ability to dynamically determine optimal timing for translation. Leveraging the advanced generation capabilities of LLMs, LSG produces translations that not only maintain semantic fidelity with the reference but also exhibit enhanced fluency and naturalness in the target language.

This case study showcases the effectiveness of our approach. Without the need for policy learning, LSG navigates the complexities of simultaneous generation, achieving the goal of low latency and high quality.

\subsection{Experiment Details}
In this section, we delineate the system settings of our LSG method. 

For the CoVoST2 Fr$\Rightarrow$En dataset, we directly leverage the Qwen-Audio \citep{chu2023qwenaudioadvancinguniversalaudio} as the LLM in our method. We set the segment size to 640 ms, and employ greedy search for inference. 

For the WMT15 De$\Rightarrow$En and MuST-C En$\Rightarrow$De datasets, we maintain consistency with the settings described in \citet{guo2024agentsimtagentassistedsimultaneousmachine}. Given the inferior generation performance of the vanilla \texttt{Llama2-7B-chat}\footnote{\url{https://huggingface.co/meta-llama/Llama-2-7b-chat-hf}}, we fine-tune the LLMs on 100k samples using LoRA \citep{hu2021lora}. For the adapters of LoRA, we set $r$ to 8. We utilize a learning rate of 0.0001 and a batch size of 128. During inference, our method implements greedy search.

To investigate the relationships between the employed LLMs and simultaneous generation performance, we conduct SimulT2TT experiments on the WMT22 Chinese$\Rightarrow$English dataset. We directly employ Bayling-7B \citep{zhang2023bayling}, Bayling-13B, and ParroT-7B \citep{jiao-etal-2023-parrot} to validate our method, utilizing greedy search for inference across all LLMs.

\begin{figure*}[t]
    \centering
    \includegraphics[width=6.0in]{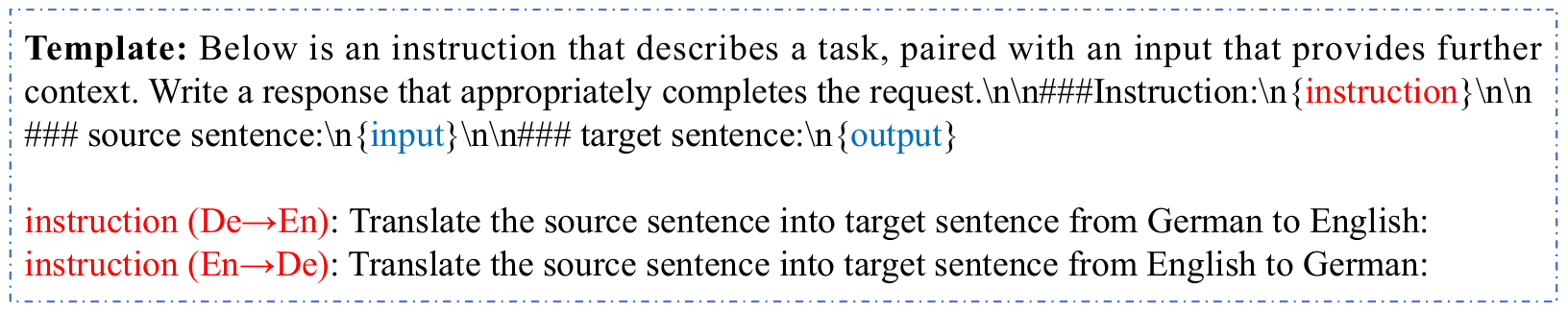}
    \caption{The prompt template of LLMs on the SimulT2TT task, where `input' denotes the partial source sentence and `output' represents the generated translation.}
    \label{t2tt_template}
\end{figure*}

\begin{figure*}[t]
    \centering
    \includegraphics[width=6.0in]{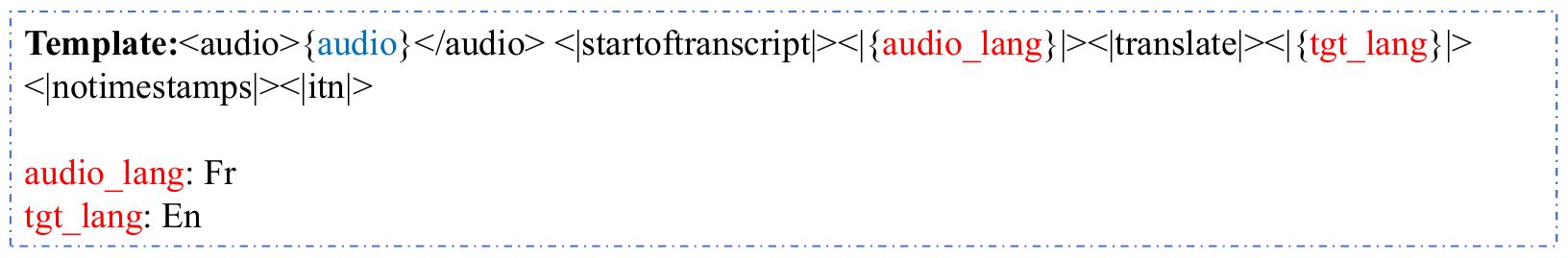}
    \caption{The prompt template of LLMs on the SimulS2TT task, where `audio' denotes the input audio.}
    \label{s2tt_template}
\end{figure*}

\begin{figure*}[t]
    \centering
    \includegraphics[width=6.0in]{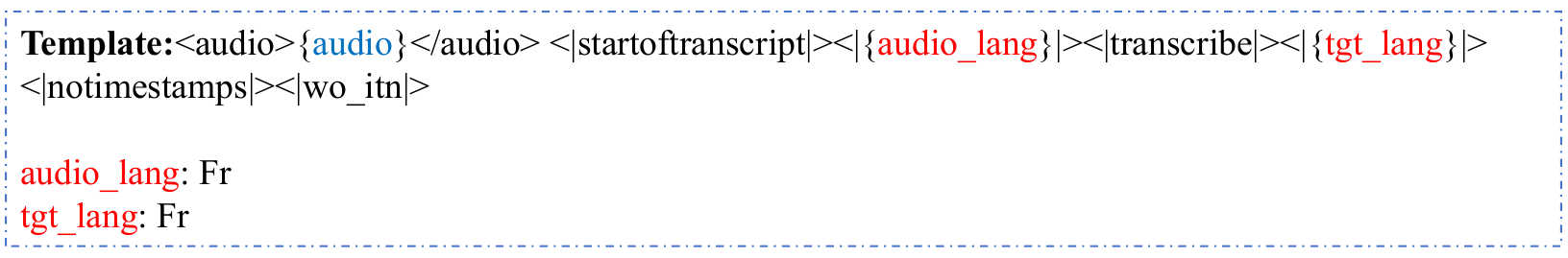}
    \caption{The prompt template of LLMs on the Streaming ASR task, where `audio' denotes the input audio.}
    \label{asr_template}
\end{figure*}

We also provide the prompt templates for LLMs on different tasks in Figures \ref{t2tt_template}, \ref{s2tt_template}, and \ref{asr_template}.

\subsection{Numerical Results}
In addition to the results presented in Figure 3, we provide the numeric outcomes of the main experiment for a more comprehensive comparison.

For the WMT15 De$\Rightarrow$En dataset, we set $\delta$ to $9.0$ and $\alpha$ to $0.6$. Table \ref{DeEn} details the range of hyperparameters and the corresponding results of LSG. Alongside the SacreBLEU metric, we also compute the COMET metric \citep{rei-etal-2020-comet} using \texttt{Unbabel/wmt22-comet-da}\footnote{\url{https://huggingface.co/Unbabel/wmt22-comet-da}} model to assess translation quality.

For the MuST-C En$\Rightarrow$De dataset, we set $\delta$ to $7.5$ and $\alpha$ to $0.6$. Table \ref{EnDe} outlines the corresponding LSG results. The COMET metric is also evaluated for this dataset.

For the CoVoST2 Fr$\Rightarrow$En dataset, we configure $\delta$ to $7.0$ and $\alpha$ to $0.5$. Table \ref{FrEn} demonstrates the range hyperparameters and corresponding results of LSG, with the COMET metric similarly assessed for this dataset.

%

\begin{table*}[]
\centering
\begin{tabular}{c  c c c} \toprule[1.2pt]
$[L, U]$ & \textbf{AL (word) ($\downarrow$)} & \textbf{SacreBLEU ($\uparrow$)} & \textbf{COMET ($\uparrow$)} \\
\cmidrule(lr){1-1} \cmidrule(lr){2-4}

$[1, 4]$ & 2.78 & 28.60 & 81.22 \\

$[3, 4]$ & 4.42 & 31.60 & 84.10 \\

$[5, 6]$ & 7.37 & 33.22 & 85.17 \\

$[7, 6]$ & 8.95 & 33.41 & 85.29\\

 \bottomrule[1pt]
\end{tabular}
\caption{The performance of LSG method on the WMT15 De$\Rightarrow$En dataset.}
\label{DeEn}
\end{table*}

\begin{table*}[]
\centering
\begin{tabular}{c  c c c} \toprule[1.2pt]
$[L, U]$ & \textbf{AL (word) ($\downarrow$)} & \textbf{SacreBLEU ($\uparrow$)} & \textbf{COMET ($\uparrow$)} \\
\cmidrule(lr){1-1} \cmidrule(lr){2-4}

$[1, 4]$ & 0.29 & 19.18 & 69.47 \\

$[3, 4]$ & 2.60 & 26.61 & 79.29 \\

$[5, 6]$ & 5.34 & 29.25 & 82.38 \\

$[7, 6]$ & 7.17 & 29.87 & 83.30 \\

 \bottomrule[1pt]
\end{tabular}
\caption{The performance of LSG method on the MuST-C En$\Rightarrow$De dataset.}
\label{EnDe}
\end{table*}

\begin{table*}[]
\centering
\begin{tabular}{c  c c c c} \toprule[1.2pt]
$[L, U]$ & \textbf{AL (ms) ($\downarrow$)} & \textbf{AL\_CA (ms) ($\downarrow$)} & \textbf{SacreBLEU ($\uparrow$)} & \textbf{COMET ($\uparrow$)} \\
\cmidrule(lr){1-1} \cmidrule(lr){2-5}

$[1, 4]$ & 1553.85 & 3318.68 & 32.10 & 77.11 \\

$[2, 4]$ & 2267.15 & 4281.70 & 34.55 & 79.43 \\

$[3, 4]$ & 3022.18 & 4550.23 & 36.19 & 80.41 \\

$[5, 6]$ & 4215.98 & 5733.66 & 37.76 & 81.76 \\

 \bottomrule[1pt]
\end{tabular}
\caption{The performance of LSG method on the CoVoST2 Fr$\Rightarrow$En dataset.}
\label{FrEn}
\end{table*}

\end{document}